\documentclass[conference]{IEEEtran}
\IEEEoverridecommandlockouts

\usepackage{cite}
\usepackage{amsmath,amssymb,amsfonts}
\usepackage{algorithmic}
\usepackage{graphicx}
\usepackage{textcomp}
\usepackage{xcolor}
\def\BibTeX{{\rm B\kern-.05em{\sc i\kern-.025em b}\kern-.08em
    T\kern-.1667em\lower.7ex\hbox{E}\kern-.125emX}}

\usepackage{multirow} 
\usepackage{booktabs} 

\begin{document}
\title{Synergistic Spotting and Recognition of Micro-Expression via Temporal State Transition
\thanks{*Corresponding author}
\thanks{This work was supported in part by the National Natural Science Foundation of China (62206015, 62227801, U20B2062), the Fundamental Research Funds for the Central Universities (FRF-TP-22-043A1), and the Young Scientist Program of The National New Energy Vehicle Technology Innovation Center (Xiamen Branch).}
}

\author{
\IEEEauthorblockN{Bochao Zou$^a$, Zizheng Guo$^a$, Wenfeng Qin$^a$, Xin Li$^b$, Kangsheng Wang$^a$ and Huimin Ma$^{a*}$}
\IEEEauthorblockA{$^a$ School of Computer and Communication Engineering, University of Science and Technology Beijing, Beijing, China}
\IEEEauthorblockA{$^b$ School of Intelligence Science and Technology, University of Science and Technology Beijing, Beijing, China}
\IEEEauthorblockA{zoubochao@ustb.edu.cn, \{guozizheng,m202220898,m202210590,m202320975\}@xs.ustb.edu.cn, mhmpub@ustb.edu.cn}
}

\maketitle

\begin{abstract}
Micro-expressions are involuntary facial movements that cannot be consciously controlled, conveying subtle cues with substantial real-world applications. The analysis of micro-expressions generally involves two main tasks: spotting micro-expression intervals in long videos and recognizing the emotions associated with these intervals. Previous deep learning methods have primarily relied on classification networks utilizing sliding windows. However, fixed window sizes and window-level hard classification introduce numerous constraints. Additionally, these methods have not fully exploited the potential of complementary pathways for spotting and recognition. In this paper, we present a novel temporal state transition architecture grounded in the state space model, which replaces conventional window-level classification with video-level regression. Furthermore, by leveraging the inherent connections between spotting and recognition tasks, we propose a synergistic strategy that enhances overall analysis performance. Extensive experiments demonstrate that our method achieves state-of-the-art performance. The codes and pre-trained models are available at https://github.com/zizheng-guo/ME-TST.
\end{abstract}


\section{Introduction}
\label{sec:intro}

Facial expressions reflect emotions and convey subtle cues such as intentions, choices, and preferences. Depending on intensity and duration, expressions can be categorized into macro-expressions (MaEs) and micro-expressions (MEs). While the emotions expressed by MaEs are not always authentic, MEs can reveal true emotions, particularly when individuals attempt to conceal or suppress their feelings. Thus, MEs are of considerable importance in areas such as criminal interrogation and business negotiations \cite{tpami-survey,tac-survey,casme3}.

ME analysis comprises two sub-tasks: spotting and recognition. Spotting refers to the precise identification of the onset and offset of MEs in long videos, whereas recognition involves classifying the emotions associated with the spotted intervals. In recent years, research efforts have primarily addressed these two tasks separately. For spotting, there has been a noticeable trend from traditional signal processing methods to deep learning approaches \cite{mesnet,lgsnet}, although traditional signal processing methods still demonstrate superior performance \cite{MEGC2023}. In contrast, research on recognition is more active \cite{recog1,recog2,recog_bert}, with deep learning approaches holding a dominant position.

\begin{figure}[t] 
  \centering 
  \setlength{\abovecaptionskip}{0.0cm}
  \includegraphics[width=0.92\linewidth]{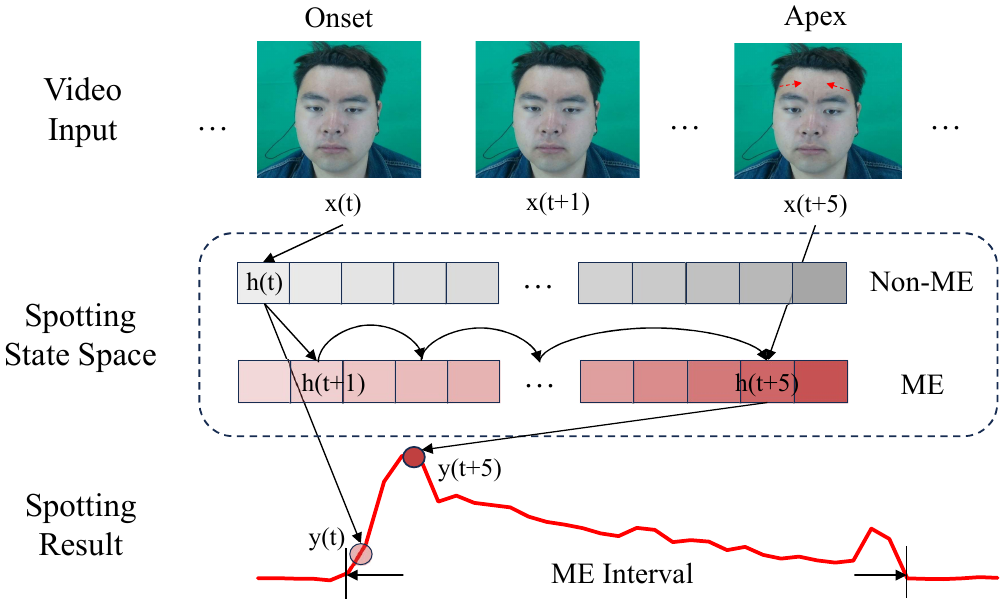}  
  \caption{A schematic diagram of state transitions. A non-ME state, $h(t)$, transitions to an ME state, $h(t+1)$, upon receiving the input from the onset frame, $x(t)$, subsequently outputting $y(t)$. As subsequent frames are processed, the ME state progressively strengthens until reaching the apex.}
  \label{fig:1}
\end{figure}

However, few studies have directly addressed the integrated ME analysis task, which is more aligned with real-world applications. Currently, attempts in the field of ME analysis remain limited. Traditional signal processing methods, as used in \cite{Li2018_MA} and \cite{Liong2017_MA}, extract features for spotting and recognition, respectively. These approaches rely on handcrafted features and are designed to analyze MEs in short videos containing a single ME, without involving MaEs. More recent methods, such as MEAN \cite{MEAN} and SFAMNet \cite{sfamnet}, utilize deep learning techniques to first spot and then recognize MEs in long videos. However, these methods are built on classification networks using sliding windows, where fixed window sizes and window-level hard classification impose constraints. These methods struggle to capture the full spectrum of ME variations, limiting the overall performance. Moreover, they do not fully exploit the potential advantages of the complementary for spotting and recognition.

\begin{figure*}[t]
  \setlength{\abovecaptionskip}{0.0cm}
  \centering  
  \includegraphics[width=0.88\linewidth]{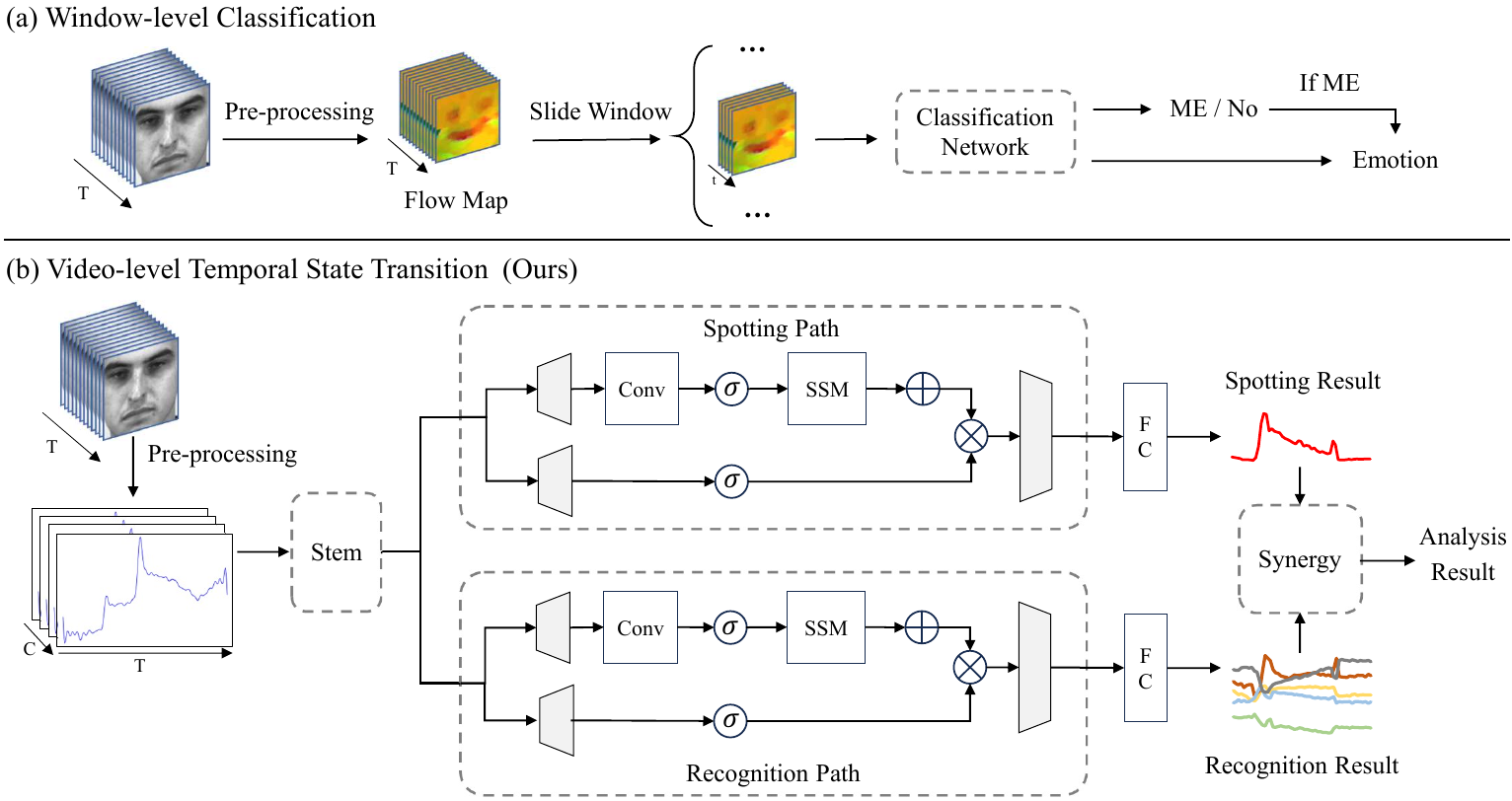}  
  \caption{(a) The framework of the window-level classification method. (b) The framework of the proposed method. Where "$+$" represents addition, "$\times$" represents multiplication, "$\sigma$" represents the activation layer, and the trapezoid represents the linear layer.}
  \label{fig:main}
\end{figure*}

In this paper, we propose a ME analysis method based on temporal state transition. As shown in Fig. \ref{fig:1}, the temporal variations in the stages of MEs, such as onset, apex, and offset, are modeled as state transitions within the state space. We replace the previous window-level hard classification with video-level regression, as illustrated in Fig. \ref{fig:main}. This architecture enables the model to handle inputs of arbitrary length during inference while maintaining linear complexity, which presents a significant advantage when processing long videos. Furthermore, we propose a synergistic spotting and recognition strategy that endows the recognition path with partial spotting capability. By leveraging the complementary advantages of both paths at the feature level and result level, this strategy effectively enhances overall performance.

The main contributions can be summarized as follows:

$\bullet$ We propose a temporal state transition architecture, which replaces window-level classification with video-level regression. To the best of our knowledge, this is the first work to investigate state space models in the ME domain.

$\bullet$ By leveraging the potential connections between spotting and recognition, we design a synergistic spotting and recognition strategy to optimize overall analysis performance.

$\bullet$ Extensive experiments demonstrate that the proposed method achieves state-of-the-art performance, with only 18K parameters and 1M multiply-accumulate operations (MACs).
\section{Methodology}
\label{sec/3_methodology} 

\subsection{The General Framework of the Proposed Method} 
The framework of the proposed method is illustrated in Fig. \ref{fig:main}b. For facial video input, the pre-processing step first extracts optical flow from the regions of interest (ROIs) \cite{MEGC2023-2, MEGC2021-1}. This produces the inter-frame optical flow sequence for the ROIs denoted as $X_{flow} \in \mathbb{R}^{C\times T}$. Here, $C$ represents the number of channels (the number of ROIs), and $T$ represents the number of video frames. Each frame is calibrated using global optical flow for the facial region.

The optical flow sequence is then processed by the stem module, which performs initial feature extraction through two 1D convolutional layers, followed by batch normalization and ReLU activation layers. The output of the stem module is fed into two pathways: one for spotting and the other for recognition, producing respective results, denoted as $X_{spot} \in \mathbb{R}^{T\times 1}$, $X_{recog} \in \mathbb{R}^{T \times (emo+1)}$. Here, $emo$ represents the number of emotion categories, with an additional category added to identify whether the expression is neutral or an ME.

Finally, the spotting and recognition results are synergized to analyze MEs in the post-processing. Specifically, peak detection is applied to the spotting results to obtain the ME intervals, and the mode of the recognition results is then used to determine the emotion corresponding to those intervals. By employing the synergy strategy, the final ME analysis result is derived, which includes the ME intervals and their corresponding emotions.

\subsection{Temporal State Transition}
Classification networks based on sliding windows are inherently constrained by fixed window size and hard classification, making it difficult to fundamentally learn the patterns of expression variations. In contrast, we approach the ME analysis as a regression task, using temporal state transitions to accurately represent the progression of ME states from onset to apex and offset. It can handle inputs of arbitrary length during the inference phase with linear complexity, making it well-suited for long videos. Specifically, we implement the Temporal State Transition based on Mamba, which can capture long-range dependencies effectively while maintaining linear complexity  \cite{mamba2,videomamba,rhythmmamba}. The channel interaction in the linear layer facilitates the exchange of information among various ROIs, allowing the network to learn the relationship between the combination of ROI movements and emotional expressions. The state space model (SSM) analyzes facial expression states from the optical flow sequences of all ROIs.

\subsection{Synergistic Spotting and Recognition}
\begin{figure}[t]
  \setlength{\abovecaptionskip}{0.00cm}
  \centering  
  \includegraphics[width=0.92\linewidth]{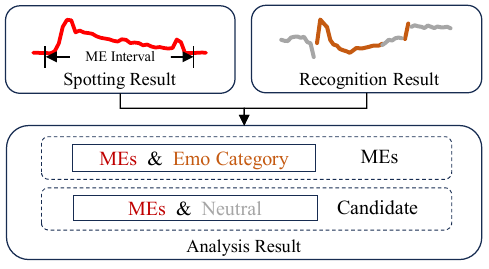}  
  \caption{The schematic diagram of the result-level synergy strategy.}
  \label{fig:2}
\end{figure}

In previous work, recognition focused solely on classifying the emotions associated with ME intervals. Therefore, the neutral category was either excluded during recognition or its loss weight was set to zero \cite{MEAN,sfamnet}. The core of the proposed Synergistic Spotting and Recognition strategy is that including the neutral category during training essentially assigns partial spotting capability to the recognition pathway. This capability facilitates synergy between the spotting and recognition pathways at both the feature and result levels: first, the spotting and recognition dual pathways share a stem module used for primary feature extraction. The spotting capability in the recognition pathway aligns partial features of spotting and recognition, thereby facilitating the learning of primary features by the stem module and enhancing overall performance. Second, recognition also performs partial spotting functions. As illustrated in Fig. \ref{fig:2}, when spotting identifies an ME but recognition classifies it as neutral, it is considered an ME candidate. 
We hypothesize that recognition is more reliable than spotting under conditions with high-motion noise, such as blinking, where the candidate is not considered an ME, and vice versa. 
If a candidate is determined to be an ME, its emotion prediction will be adjusted to the emotion with the highest probability in the distribution, excluding neutral.

\label{sec/4_experiment}
\section{Experiment}

\subsection{Dataset and Performance Metric}
The CAS(ME)$^3$ \cite{casme3}  and SAMMLV \cite{samm,SAMMLV} long video datasets were used for evaluation. CAS(ME)$^3$ includes 1300 long videos from 100 subjects, containing 3342 MaEs and 860 MEs, with a frame rate of 30 fps. SAMMLV includes 147 long videos from 32 subjects, containing 343 MaEs and 159 MEs, only MEs have emotion labels, with a frame rate of 200 fps.
The Spot-Then-Recognize Score (STRS) is used to evaluate ME analysis \cite{MEAN}. The F1 score and Recall assess ME spotting, while the F1 Score, unweighted average recall (UAR) and unweighted F1-score (UF1) evaluate recognition \cite{uf1uar}.
For spotting, we follow the protocol in \cite{iou}, where an Intersection over Union greater than 0.5 between the ground-truth interval and the spotted interval is considered a True Positive.

\subsection{Implementation Details}
We implemented the proposed method using PyTorch. Optical flow computation was performed using the Gunnar Farneback algorithm \cite{flow}. The loss function for spotting is the Mean Squared Error (MSE), while the loss function for recognition is cross-entropy. The Leave-One-Subject-Out cross-validation protocol is used for evaluation. Adam optimizer was used, and the learning rate scheduler followed the 1cycle policy. The model was trained for 30 epochs, with the loss function set to 1e-4. The network training was performed on an RTX 4090.

\subsection{ME Spotting Evaluation}
As shown in Table \ref{tab:mes}, following the protocol in \cite{casme3,sfamnet}, we compared the spotting performance of the proposed method with previous approaches, including those of type S (spotting only) and S\&R (spotting and recognition simultaneously). It can be observed that the proposed method achieved state-of-the-art performance in ME spotting, outperforming not only the methods designed for both spotting and recognition but also those dedicated solely to spotting.

\begin{table}[t]
  \centering
  \setlength{\abovecaptionskip}{0.05cm}
  \setlength{\tabcolsep}{6.5pt}
  \caption{ME Spotting Evaluation.}
    \begin{tabular}{ccccc}
    \toprule
    \multirow{2}[4]{*}{Type} & \multirow{2}[4]{*}{Method} & CAS(ME)$^3$ & \multicolumn{2}{c}{SAMMLV} \\
\cmidrule{3-5}     &     & F1 Score$\uparrow$ & Recall$\uparrow$ & F1 Score$\uparrow$\\
    \midrule
    \multirow{7}[2]{*}{S} & He et al. \cite{he}  & -     & 0.0360 & 0.0364 \\
          & SP-FD \cite{zhang2020} & 0.0103 & 0.1330 & 0.1331 \\
          & OF-FD \cite{MEGC2021-1} & 0.0000 & 0.2160 & 0.2162 \\
          & MESNet \cite{mesnet} & - & - & 0.0880 \\
          & Yap et al. \cite{yap}  & - & 0.0440 & - \\
          & Liong et al. \cite{liong2021}  & - & - & 0.1520 \\
          & LSSNet \cite{lssnet} & 0.0653 & 0.2120 & 0.1310 \\
          & LGSNet \cite{lgsnet} & - & 0.2570 & - \\
    \midrule
    \multirow{3}[2]{*}{S \& R} & MEAN \cite{MEAN}  & 0.0283 & 0.1635 & 0.0949 \\
          & SFAMNet \cite{sfamnet}& 0.0716 & -     & - \\
          & Ours  & \textbf{0.0802} & \textbf{0.3019} & \textbf{0.2167} \\
    \bottomrule
    \end{tabular}%
  \label{tab:mes}%
\end{table}%

\begin{table}[t]
  \vspace{-0.2cm}
  \centering
  \setlength{\abovecaptionskip}{0.05cm}
  \setlength{\tabcolsep}{15.0pt}
  \caption{ME Recognition Evaluation.}
    \begin{tabular}{cccc}
    \toprule
    Type & Method & UF1$\uparrow$  & UAR$\uparrow$\\
    \midrule
    \multirow{4}[2]{*}{R} & STSTNet \cite{ststnet} & 0.3795 & 0.3792 \\
          & RCN-A \cite{RCN-A} & 0.3928 & 0.3893 \\
          & FeatRef \cite{FR} & 0.3493 & 0.3413 \\
          & AlexNet \cite{casme3} & 0.3001 & 0.2982 \\
    \midrule
    \multirow{3}[2]{*}{S \& R} & MEAN \cite{MEAN} & 0.3894 & 0.4004 \\
          & SFAMNet \cite{sfamnet} & 0.4462 & 0.4767 \\
          & Ours  & \textbf{0.4754} & \textbf{0.4878} \\
    \bottomrule
    \end{tabular}%
  \label{tab:mer}%
\end{table}%

\begin{figure*}[t]
  \setlength{\abovecaptionskip}{0.0cm}
  \centering  
  \includegraphics[width=0.92\linewidth]{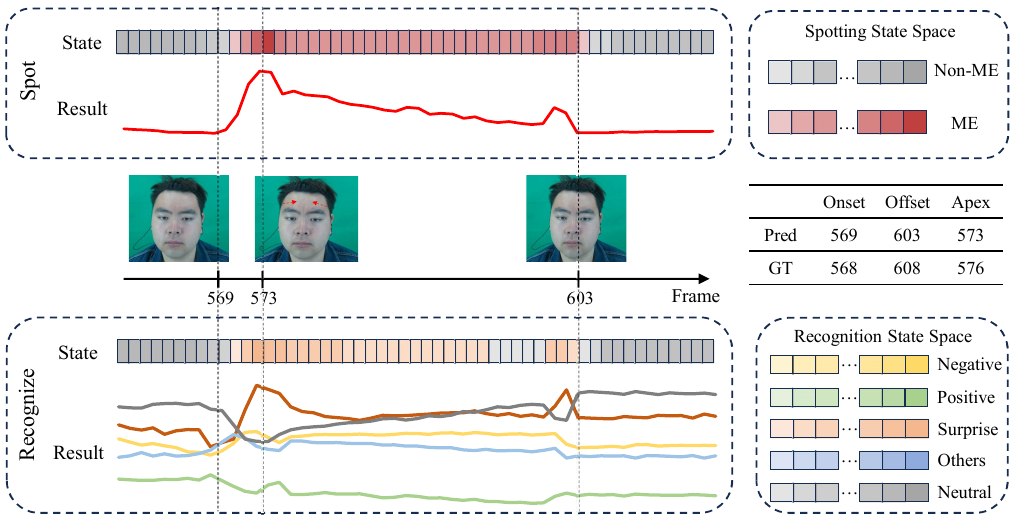}  
  \caption{Visualization of an example from the CAS(ME)$^3$ dataset results.}
  \label{fig:vis}
\end{figure*}

\begin{table}[t]
  \centering
  \setlength{\abovecaptionskip}{0.05cm}
  \setlength{\tabcolsep}{4.0pt}
  \caption{ME Analysis Evaluation. \\Analysis: STRS$\uparrow$; Spotting and Recognition: F1 Score$\uparrow$.}
    \begin{tabular}{ccccccc}
    \toprule
    \multirow{2}[4]{*}{Method} & \multicolumn{3}{c}{CAS(ME)$^3$} & \multicolumn{3}{c}{SAMMLV} \\
    \cmidrule{2-7}   & Analy & Spot  & Recog & Analy & Spot  & Recog \\
    \midrule
    MEAN \cite{MEAN} & 0.0100 & 0.0283 & 0.3532 & 0.0499 & 0.0949 & 0.5263 \\
    SFAMNet \cite{sfamnet} & 0.0331 & 0.0716 & 0.4619 & -     & -     & - \\
    Ours  & \textbf{0.0387} & \textbf{0.0802} & \textbf{0.4830} & \textbf{0.1356} & \textbf{0.2167} & \textbf{0.6259} \\
    \bottomrule
    \end{tabular}%
  \label{tab:mea}%
\end{table}%

\begin{table}[t]
  \vspace{-0.2cm}
  \centering
  \setlength{\abovecaptionskip}{0.05cm}
  \caption{Ablation study of synergy. \\Analysis: STRS$\uparrow$; Spotting and Recognition: F1 Score$\uparrow$.}
    \begin{tabular}{ccccccc}
    \toprule
    \multirow{2}[4]{*}{Synergy} & \multicolumn{3}{c}{CAS(ME)$^3$} & \multicolumn{3}{c}{SAMMLV} \\
\cmidrule{2-7}          & Analy & Spot  & Recog & Analy & Spot  & Recog \\
    \midrule
    w/o   & 0.0314 & 0.0753 & 0.4171 & 0.1321 & 0.2130 & 0.6202 \\
    w.    & \textbf{0.0387} & \textbf{0.0802} & \textbf{0.4830} & \textbf{0.1356} & \textbf{0.2167} & \textbf{0.6259} \\
    \bottomrule
    \end{tabular}%
  \label{tab:ablation}%
\end{table}%

\subsection{ME Recognition Evaluation}
As shown in Table \ref{tab:mer}, we conducted ME recognition evaluation on CAS(ME)$^3$ dataset, following the protocol in \cite{casme3,sfamnet} where MEs are classified into four categories. It can be seen that the proposed method not only demonstrates excellent spotting performance but also achieves state-of-the-art recognition performance.

\subsection{ME Analysis Evaluation}
We conducted the ME analysis evaluation to comprehensively validate the effectiveness of the proposed method. For the CAS(ME)$^3$ dataset, MEs were categorized into four categories~\cite{sfamnet}. As to the SAMMLV dataset, only three categories (negative, surprise, and positive) were recognized~\cite{MEAN}. As shown in Table \ref{tab:mea}, the proposed method achieved state-of-the-art performance compared to previous approaches, thoroughly demonstrating the feasibility of the video-level regression architecture and highlighting its advantages over window-level classification architecture.

\subsection{Ablation Study}
We conducted the ablation study to assess the impact of the synergy strategy. In the comparison experiments, the loss weight for the neutral category was set to 0. As shown in Table \ref{tab:ablation}, the synergy strategy effectively improves overall analysis performance. This indicates that the proposed synergy strategy effectively leverages the complementary advantages of both spotting and recognition, resulting in a synergistic effect where the combined performance surpasses the mere sum of the individual contributions.

\subsection{Visualization}
We visualize an example of the CAS(ME)$^3$ results in Fig. \ref{fig:vis}. At key points such as onset, offset, and apex, both spotting and recognition results exhibit distinct responses, indicating that the proposed method effectively characterizes ME states and validates the feasibility of the temporal state transition architecture. Particularly, the comparison between the surprise emotion curve (orange) within the recognition results and the spotting results highlights that the recognition path exhibits a certain degree of spotting capability. 
\label{sec/5_conclusion}
\section{Conclusion}

In this paper, we propose a novel temporal state transition architecture, which transforms window-level hard classification into video-level regression with a low computational cost of 1.370M MACs and 18.054K parameters. This architecture better aligns with the nature of MEs, and its linear complexity enables it to efficiently handle long videos. Additionally, we leverage the potential connections between spotting and recognition to propose a synergy strategy, which optimizes the overall performance of ME analysis. In future work, we plan to further investigate the connection between the two at the state level and attempt to address the ME analysis task through a single recognition pathway.

\bibliographystyle{IEEEtran}
\bibliography{ref}

\end{document}